\documentclass[lettersize,journal]{IEEEtran}
\usepackage{amsmath,amsfonts}
\usepackage{algorithmic}
\usepackage{algorithm}
\usepackage{array}
\usepackage[caption=false,font=normalsize,labelfont=sf,textfont=sf]{subfig}
\usepackage{textcomp}
\usepackage{stfloats}
\usepackage{url}
\usepackage{verbatim}
\usepackage{graphicx}
\usepackage{cite}
\usepackage{booktabs}
\usepackage{multirow}
\usepackage{bbding}
\usepackage{hyperref}
\usepackage {xcolor}
\usepackage{colortbl}
\usepackage[flushleft]{threeparttable}
\begin{document}

\title{\textit{FeatAug-DETR}: Enriching One-to-Many Matching for DETRs with Feature Augmentation}

\author{Rongyao~Fang, Peng~Gao, Aojun~Zhou, Yingjie~Cai, Si Liu, Jifeng Dai, and~Hongsheng~Li \thanks{R. Fang, A. Zhou, Y. Cai, H. Li are with the Chinese University of Hong Kong, Hong Kong SAR, China.} \thanks{P. Gao is with Shanghai AI Laboratory, Shanghai, China.} \thanks{S. Liu is with Beihang University, Beijing, China.} \thanks{J. Dai is with Tsinghua University, Beijing, China.}\thanks{Corresponding author: Hongsheng Li}}

\maketitle

\begin{abstract}
One-to-one matching is a crucial design in DETR-like object detection frameworks. It enables the DETR to perform end-to-end detection. However, it also faces challenges of lacking positive sample supervision and slow convergence speed. Several recent works proposed the one-to-many matching mechanism to accelerate training and boost detection performance. We revisit these methods and  model them in a unified format of augmenting the object queries. In this paper, we propose two methods that realize one-to-many matching from a different perspective of augmenting images or image features. The first method is One-to-many Matching via Data Augmentation (denoted as \textit{DataAug-DETR}). It spatially transforms the images and includes multiple augmented versions of each image in the same training batch. Such a simple augmentation strategy already achieves one-to-many matching and surprisingly improves DETR’s performance. The second method is One-to-many matching via Feature Augmentation (denoted as \textit{FeatAug-DETR}). Unlike \textit{DataAug-DETR}, it augments the image features instead of the original images and includes multiple augmented features in the same batch to realize one-to-many matching. \textit{FeatAug-DETR} significantly accelerates DETR training and boosts detection performance while keeping the inference speed unchanged. We conduct extensive experiments to evaluate the effectiveness of the proposed approach on DETR variants, including DAB-DETR, Deformable-DETR, and $\mathcal{H}$-Deformable-DETR. Without extra training data, \textit{FeatAug-DETR} shortens the training convergence periods of Deformable-DETR \cite{2021deformable_detr} to 24 epochs and achieves 58.3 AP on COCO \texttt{val2017} set with Swin-L as the backbone. Code
will be available at: \href{https://github.com/rongyaofang/FeatAug-DETR}{https://github.com/rongyaofang/FeatAug-DETR}
\end{abstract}


\section{Introduction}
\label{sec:intro}
Object detection is a fundamental task in computer vision, which predicts bounding boxes and categories of objects in an image. In the past several years, deep learning made significant success in the object detection task and tens of classic object detectors have been proposed. These classic detectors are mainly based on convolutional neural networks, which include the one-stage detectors \cite{2016yolo,2017yolov2,2018yolov3} and two-stage detectors \cite{2014rcnn,2015fasterrcnn1,2017fasterrcnn2,2017maskrcnn}. One-to-many label assignment is the core design of the classic detectors, where each ground-truth box is assigned to multiple predictions of  the detector. With such a one-to-many matching scheme, these frameworks require human-designed non-maximum suppression (NMS) for post-processing and cannot be trained end-to-end.

\begin{figure}[tb!]
   \centering
   \includegraphics[width=0.95\linewidth]{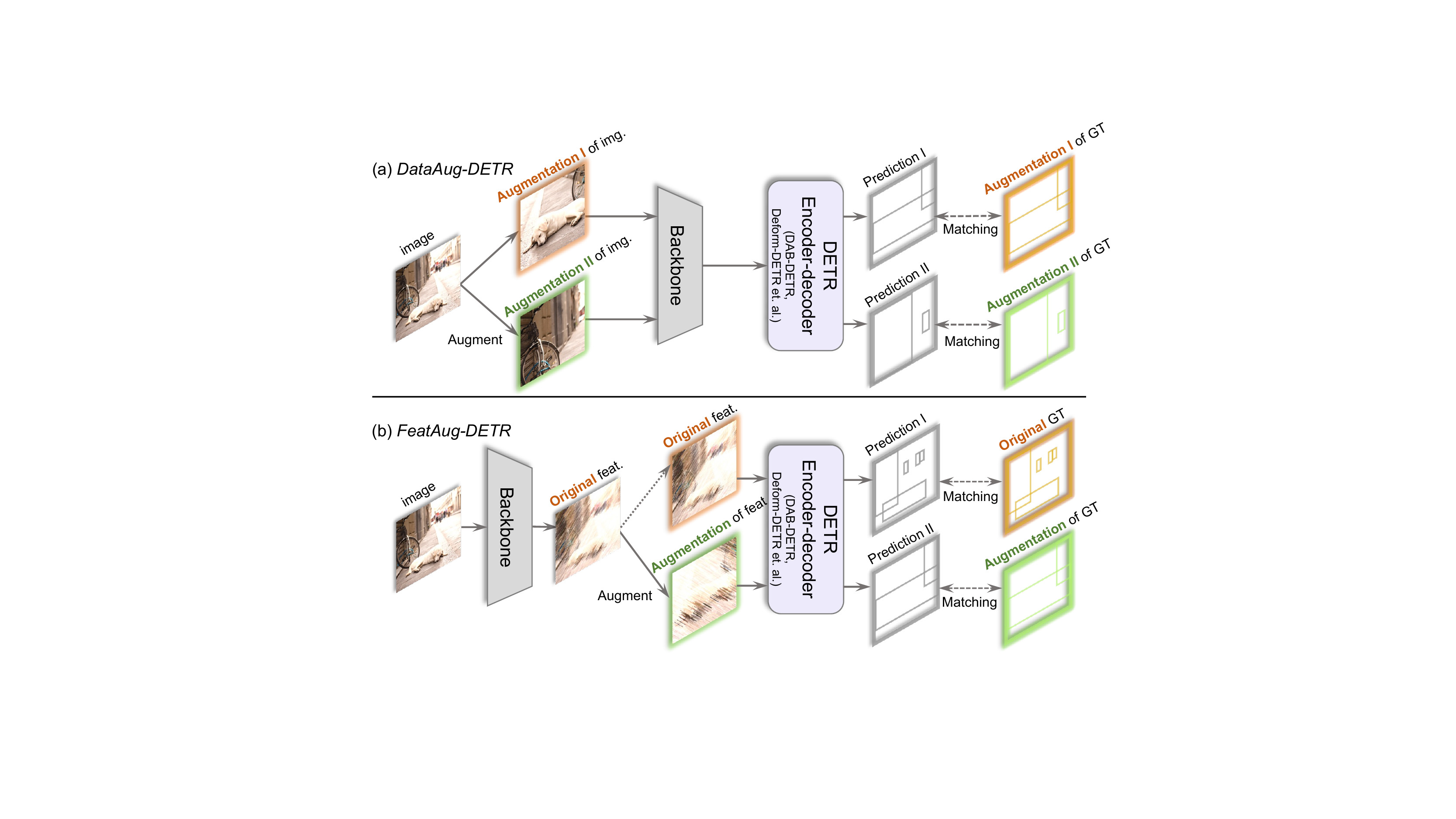}

   \caption{Previous works achieve one-to-many matching via augmenting object queries \cite{2022group_detr,2022hybrdi_matching}. We propose \textit{DataAug-DETR} and \textit{FeatAug-DETR} to achieve one-to-many matching from the perspective of augmenting image features and ground truth objects.}
   \label{fig:intro}
\end{figure}

The DEtection TRansformer (DETR) \cite{2020detr} reveals the potential of using transformer architectures \cite{2016transformer} to achieve high performance in object detection. It implements one-to-one matching between the ground truth boxes and the predictions. Various DETR-like frameworks have shown great success. Quite a few follow-up works were proposed to improve DETR by modifying the architectures, such as the architecture of the transformer encoder \cite{2021deformable_detr,2022cf_detr}, the decoder \cite{2021deformable_detr,2022dn_detr,2021smca,2021dynamic_detr}, and the query designs \cite{2022anchor_detr,2021conditional_detr,2022dab_detr}.

Besides improving the DETR architectures, several recent works aim to improve the one-to-one matching mechanism in DETR \cite{2020detr}. The one-to-one matching helps DETR to discard the human-designed NMS for post-processing. In addition, \cite{2022group_detr} showed that extra one-to-many matching supervisions can lead to faster and better convergence. In the recent Group-DETR \cite{2022group_detr} and Hybrid Matching \cite{2022hybrdi_matching}, extra object queries are used to formulate one-to-many matchings to provide additional supervisions for better training DETRs.

The one-to-many matching mechanism associates each ground truth object with multiple object queries. The object queries interact with the image feature maps containing the objects via cross-attention in the DETR decoder. The one-to-one and one-to-many matchings therefore implicitly conduct assignments between queries and the object features from the spatial feature maps. State-of-the-art Group-DETR and Hybrid Matching enhance one-to-one matching by augmenting extra object queries and inputting them into the matching module. They achieve impressive results in accelerating convergence speed and boosting detection performance.

Our initial observation highlights that a straightforward yet appropriately designed data augmentation scheme (\textit{DataAug-DETR}) can implicitly accomplish one-to-many matching and surprisingly improve DETR performance. By integrating numerous spatially augmented versions of a single image in a single batch, the same objects can be assigned to distinct queries across various augmented images. These one-to-many assignments can greatly enhance detection performance.

Given that DETR queries accumulate information from image feature maps, we propose approximating the impact of spatial image augmentation by applying spatial augmentation to the feature maps, thereby avoiding repeated forwarding of different versions of the same image into the vision backbone. We further propose feature augmentation (\textit{FeatAug-DETR}) for DETR, which spatially shifts and flips feature maps and arranges different versions of a feature map in the same batch, thereby assigning the same object queries to different objects after feature augmentation. This method is a simple yet effective way to enhance DETR performance.

We conduct extensive experiments to evaluate the efficiency and effectiveness of \textit{DataAug-DETR} and \textit{FeatAug-DETR}. As a plug-and-play approach, our proposed modules can be easily integrated into different DETR variants. \textit{FeatAug-DETR} significantly accelerates convergence and also improves the performance of various DETR detectors, including DAB-DETR \cite{2022dab_detr}, Deformable-DETR \cite{2021deformable_detr}, and $\mathcal{H}$-Deformable-DETR \cite{2022hybrdi_matching}. \textit{FeatAug-DETR} helps Deformable-DETR \cite{2021deformable_detr} with Swin-Large backbone \cite{2021swin} reach $55.8$ AP with $1\times$ training schedule (12 epochs), which is $1.3$ AP higher than that without our method ($54.5$ AP), while keeping the inference FLOPs unchanged. Moreover, \textit{FeatAug-DETR} is compatible with $\mathcal{H}$-Deformable-DETR \cite{2022hybrdi_matching} as the feature augmentation realizes one-to-many matchings from a different perspective.

In summary, our contributions are summarized as follows:

\begin{itemize}
    \item We propose \textit{DataAug-DETR}, which augments each image multiple times and includes the different augmented versions of an image in the same training batch. It boosts DETR's detection accuracy.

    \item We further propose feature augmentation (\textit{FeatAug-DETR}) for DETR. It augments the feature maps from the vision backbone and significantly accelerates training compared with \textit{DataAug-DETR}.

    \item When integrating \textit{FeatAug-DETR} into Hybrid Matching \cite{2022hybrdi_matching}, our method achieves 58.3 AP on COCO \texttt{val2017} with Swin-L backbone and 24 training epochs, surpassing the state-of-the-art performance by 0.5 AP.
\end{itemize}

\section{Related Work}
\subsection{Classic Object Detectors}
Modern object detection models are divided into 2 categories, one-stage detectors, and two-stage detectors. The one-stage detectors \cite{2017yolov2,2018yolov3} predict the positions of the objects relying on the anchors. The two-stage detectors \cite{2015fasterrcnn1,2017fasterrcnn2} first generate region proposals and then predict the object position w.r.t. the proposals. These methods are both anchor-based methods in which the predefined anchors play an important role in the models. These classic object detectors also need hand-designed operations such as NMS as post-processing, which makes they cannot optimize end-to-end.

\subsection{Label Assignment in Classic Object Detectors}
Assignment between the ground truth objects and training samples is a widely-investigated topic in classic object detectors \cite{2016yolo,2017yolov2,2015fasterrcnn1,2017maskrcnn,2017focal}. Anchor-based detectors \cite{2017focal,2016ssd} utilize Intersection-over-Union (IoU) to apply label assignment. The anchor will be assigned to the maximum IoU ground truth box when the maximum IoU between an anchor and all gt boxes exceeds the predefined IoU threshold. Anchor-free detectors \cite{2019fcos,2020atss} utilize spatial and scale constraints when selecting positive points. The follow-up works \cite{2020prob_assign,2021ota} propose improvements in a similar direction. The above label assignment methods in classic object detectors are one-to-many matching, which always assigns several object predictions with one ground truth box. Such methods require NMS for post-processing, which makes the detectors hard to train in an end-to-end manner.

\subsection{Detection Transformer}
Carion \textit{et al.} \cite{2020detr} proposed Detection Transformer (DETR), which introduces the Transformer architecture into the object detection field. They also use bipartite matching to implement set-based loss and make the framework an end-to-end architecture. These designs remove the handcraft components such as anchors and NMS in the previous classic object detectors. However, DETR still faces the problem of slow convergence speed and relatively low performance. Also, DETR only uses one scale image feature, which lost the benefit of the multi-scale feature which has been proven effective in previous works. The follow-up works proposed improvements to relieve these problems effectively. \cite{2021detr_encoder} designed an encoder-only DETR without using a decoder. Anchor-DETR \cite{2022anchor_detr} utilizes designed anchor architecture in the decoder to help accelerate training. Conditional-DETR \cite{2021conditional_detr} learns conditional spatial query to help cross-attention head to attend to a band containing a distinct region. It significantly shortens the training epochs of DETR from 500 epochs to 50 or 108 epochs. Efficient DETR \cite{2021efficient_detr} selects top $K$ positions from the encoder’s prediction to enhance decoder queries. Dynamic Head \cite{2021dynamic_head} proposed a dynamic decoder to focus on important regions from multiple feature levels. Deformable-DETR \cite{2021deformable_detr,2017deformable_conv,2019deformable_conv_v2} proposed deformable attention and replace the original attention mechanism in DETR. It makes utilizing multi-scale image feature feasible in DETR architecture. The follow-up DAB-DETR \cite{2022dab_detr} uses 4-D box coordinates as queries and updates boxes layer-by-layer in the Transformer decoder part. In the later work, DN-DETR \cite{2022dn_detr} and DINO \cite{2022dino}, they use bounding box denoising operation and continue to shorten the training period to 3$\times$ standard training schedule (36 epochs).

\subsection{One-to-one Matching}
Carion \textit{et al.} \cite{2020detr} uses bipartite matching to implement one-to-one matching between the object queries and ground truth bounding box. The pipeline computes a pair-wise matching cost depending on the class prediction and the similarity of predicted and ground truth boxes. Then the Hungarian algorithm \cite{2010hungarian} is applied to find the optimal assignment. The design helps DETR achieve end-to-end training in the object detection field without implementing the hand-designed NMS operations. 

However, because of the relatively small number of ground truth boxes in each image (always smaller than 50) compared with the number of predictions which are generally larger than 300. The positive sample supervision provided by one-to-one matching is relatively sparse \cite{2022group_detr}. It results in a slow convergence speed for the DETR framework. This problem can be effectively revealed by designing one-to-many matching methods.

\subsection{One-to-Many Matching}
Recently, several works \cite{2022group_detr,2022hybrdi_matching} discuss the sparsity positive supervision problem of one-to-one matching and proposed one-to-many matching methods on DETR-liked frameworks. Group-DETR \cite{2022group_detr} decouples the positives into multiple independent groups and keeps only one positive per gt object in each group. Hybrid Matching \cite{2022hybrdi_matching} combines the original one-to-one matching branch with auxiliary queries that use one-to-many matching loss during training. The principles of these two methods are similar. They provide extra object queries in the decoder to produce more positive supervision in the model. The two methods both effectively accelerate convergence speed and boost performance for DETR. 

Our method shares the same principle of generating more positive supervision. But unlike Group-DETR or Hybrid Matching, which implement extra object queries, we augment the images in a batch or the features from the backbone to realize the goal. Our method achieves obvious performance improvement in DETR training. Through experiments, we show that our method continues to obtain further performance boosting when applying previous methods (such as Hybrid Matching) together.

\section{Method}
\subsection{A Brief Revisit of DETR and One-to-Many Matching} \label{sec:motivation}
In DETR, an input image $I$ is processed by the backbone $\mathcal{B}$ to obtain the feature map $F\in\mathbb{R}^{\hat{C}\times \hat{H}\times \hat{W}}$. A stack of self-attentions and feed-forward networks in the DETR encoder transform to obtain the feature maps. The transformer decoder utilizes cross-attention to guide a series of object queries $Q\in\mathbb{R}^{N\times C}$ to aggregate information from the feature maps and generate object predictions $P$, where $N$ denotes the number of object queries and $C$ is the query vector dimension. The predictions include the normalized location predictions $P_{l}\in\mathbb{R}^{N\times 4}$ and class predictions $P_{c}\in\mathbb{R}^{N\times (N_{cls}+1)}$, in which $N_{cls}$ denotes the class number. The one-to-one matching strategy is adopted to associate the predictions $P$ with ground truth objects $G$.

An object query set $Q=\{q_1, q_2, \dots,q_N\}\in\mathbb{R}^{N\times C}$ is input into the Transformer decoder. The queries aggregate information from image features $F$ through the cross-attention operations in the Transformer decoder $\mathcal{D}(\cdot)$ which has $L$ Transformer layers and outputs object predictions. The predictions of each decoder layer are denoted as $P^1, P^2, \dots, P^L$ respectively. A bipartite one-to-one matching between the object predictions and the ground truth $G$ is conducted at each layer. The process is formulated as:

\begin{equation}\label{eq:one2one}
P^l=\mathcal{D}^l(Q, F), \quad
\mathcal{L}_{\mathrm{one2one}}=\sum_{l=1}^L\mathcal{L}_{\mathrm{Hungarian}}(P^l,G),
\end{equation} 
where $\mathcal{D}^l(\cdot)$ denotes the $l^{th}$ layer of the decoder, $\mathcal{L}_{\mathrm{one2one}}$ denotes the one-to-one matching loss, and $\mathcal{L}_{\mathrm{Hungarian}}$ denotes the Hungarian matching (bipartite matching) loss at each layer. Since each prediction $p_i^l\in P^l$ is generated from the object query $q_i$, the matchings between ground truths and predictions can be viewed as \textit{the matchings between ground truths $G$ and object queries $Q$}.

In order to better supervise DETR and accelerate its training, one-to-many matching schemes were proposed in \cite{2022group_detr,2022hybrdi_matching}. Group DETR \cite{2022group_detr} and Hybrid Matching \cite{2022hybrdi_matching} to augment extra object queries $\hat{Q}$ and introduce one-to-many matching loss, which achieves significant performance gain. Similar to the formulation of one-to-one matching, the general formulation of one-to-many matching can be defined as:

\begin{equation}\label{eq:one2many}
\hat{P}^l_k=\mathcal{D}^l(\hat{Q}_k, F),\ \mathcal{L}_{\mathrm{one2many}}=\sum_{k=1}^K\sum_{l=1}^L\mathcal{L}_{\mathrm{Hungarian}}(\hat{P}^l_k,\hat{G}_k),
\end{equation}

\noindent where $K$ denotes the number of Hungarian matching groups. In each matching group, the predictions corresponding to $k$-th group of augmented queries $\hat{Q}_k$ are denoted as $\hat{P}^l_k$, and the augmented ground truths are denoted as $\hat{G}_k$.

In one-to-many matching, the predictions and ground truths are augmented to generate different groups of positive supervision. Here we discuss the designs of Group-DETR \cite{2022group_detr} and Hybrid Matching \cite{2022hybrdi_matching} following the above unified formulation (Eq. (\ref{eq:one2many})).

\subsubsection{Group-DETR} 
Group-DETR utilizes $K$ separate groups of object queries $Q_1, \dots, Q_K$ and generates $K$ groups of predictions for each training image. The same set of ground truth $G$ applies one-to-one matching to each group of the predictions respectively. The process can be formulated as:

\begin{equation}\label{eq:group_detr}
P^l_k=\mathcal{D}^l(Q_k, F),\ \mathcal{L}_\mathrm{Group-DETR}=\sum_{k=1}^K\sum_{l=1}^L\mathcal{L}_{\mathrm{Hungarian}}(P^l_k,G),
\end{equation}

In Group-DETR, multiple groups of object queries are matched to the same set of ground truths. The augmentation on the object queries helps to boost the model performance.

\begin{figure*}[tb!]
  \centering
   \includegraphics[width=0.9\linewidth]{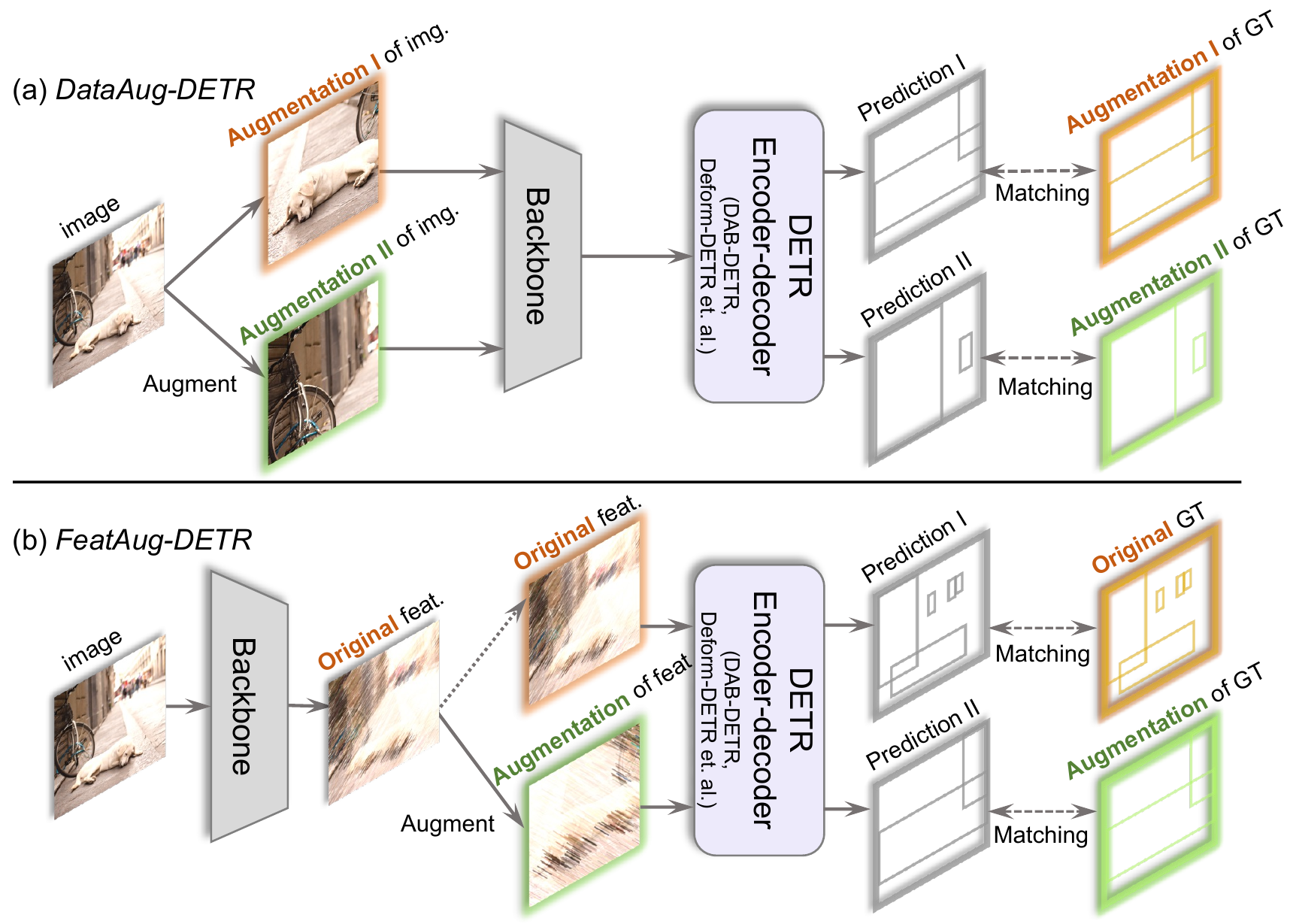}
   \caption{\textit{DataAug-DETR} augments images several times and includes multiple augmented versions in the same batch. \textit{FeatAug-DETR} augments feature maps from the vision backbone multiple times and include them in the same batch. Our \textit{FeatAug-DETR} can augment both single-scale and multi-scale feature maps. Only single-scale feature maps are shown here for simplicity.}
   \label{fig:method}
\end{figure*}

\subsubsection{Hybrid Matching}
In Hybrid Matching, it uses a second group of object queries, which contains $T$ object queries $\hat{Q}=\{\hat{q}_1, \hat{q}_2, \dots,\hat{q}_T\}$. The second group of queries applies one-to-many matching with repeated sets of ground truths $\hat{G}=\{\hat{G}_1, \hat{G}_2, \dots, \hat{G}_K\}$, where $\hat{G}_1=\hat{G}_2=\dots=\hat{G}_K=G$. The process of Hybrid Matching can be formulated as:

\begin{equation}\label{eq:hybrid_matching}
\begin{split}
&P^l=\mathcal{D}^l(Q, F), \quad \hat{P}^l=\mathcal{D}^l(\hat{Q},F), \\
&\mathcal{L}_{\mathcal{H}}=\sum_{l=1}^L\mathcal{L}_{\text{Hungarian}}(P^l,G)+\sum_{l=1}^L\mathcal{L}_{\text{Hungarian}}(\hat{P}^l,\hat{G}).
\end{split}
\end{equation}

Group-DETR and Hybrid Matching both augment the object queries $\hat{Q}$ to facilitate the training. Considering the unified formulation of one-to-many matching (Eq. (\ref{eq:one2many})), Group-DETR and Hybrid Matching both conduct one-to-many matching via augmenting the query set $Q$ but ignore the possibility of jointly augmenting the image features $F$ and the ground truths $G$.

\subsection{One-to-many Matching via Data Augmentation} \label{sec:pilot}
Our important observation is that one-to-many matching can also be implemented via augmenting the image feature $F$ and the ground truths $G$ in Eq. (\ref{eq:one2many}). We explore conducting spatial augmentation on each image multiple times and include them in the same batch. And we experimentally validate that spatially augmented versions of the same image lead to different query-ground truth assignments.

We conduct a pilot study on COCO \texttt{train2017} dataset, where we augment every image two times with random flipping and cropping. The random flipping and cropping operations follow the same operations as that in DETR \cite{2020detr} for data augmentation. Such augmented image pairs are then input to a trained Deformable-DETR whose parameters are fixed. The Deformable-DETR has 300 object queries and operates in a one-stage manner. We observe that 95.9\% of the corresponding objects in the two augmented images are assigned to different object queries. The remaining 4.1\% objects that are assigned to the same object queries are mostly located at the same relative positions in the two augmented images. More specifically, ground truth bounding boxes with unchanged queries of the two augmented images have a high average IoU of 78.2\%. This pilot study shows that, by spatially augmenting each training image in a proper way, the objects can be assigned to different object queries via spatial image augmentation. It is therefore reasonable to take advantage of image spatial transformation to modify $F$ and $G$ to achieve one-to-many matching for DETRs.

The above pilot study shows that one-to-many matching can be achieved via spatial data augmentation. We propose \textit{DataAug-DETR}, which conducts spatial image augmentation on each training image and includes them in the same training batch. We adopt the default data augmentation scheme of DETR and Deformable-DETR, which includes a $50\%$ random horizontal flipping and a $50\%$ random cropping, followed by a random resizing to several pre-defined sizes with the short edge ranging from $[480,\ 800]$. Assume that each image is spatially augmented for $N$ times in each training iteration. We denote the data augmentation operation as $\mathcal{T}_n(\cdot)$, where $n$ denotes the $n$-th random data augmentation to an image. The $N$ versions of the image pass through the image feature backbone $\mathcal{F}$ in DETR and generate $N$ image features $\{\hat{F}_n=\mathcal{F}(\mathcal{T}_n(I))\}_{n=1}^{N}$. Note that data augmentation is also applied to the ground truth labels $G$ and generates $N$ versions of labels $\{\hat{G}_n=\mathcal{T}_n(G)\}_{n=1}^{N}$. The augmentation process can be formulated as:

\begin{equation}\label{eq:data_aug}
\hat{F}_n=\mathcal{F}(\mathcal{T}_n(I)),\ 
\hat{G}_n=\mathcal{T}_n(G),\ \mathrm{for} \ n=1,\dots,N.
\end{equation}

Then by applying the bipartite matching on each augmented image individually, the matching process in Eq. (\ref{eq:one2many}) becomes:

\begin{equation}\label{eq:data_aug2}
\hat{P}^l_n=\mathcal{D}^l(Q, \hat{F}_n),\ 
\mathcal{L}_{\mathrm{DataAug}}=\sum_{n=1}^N\sum_{l=1}^L\mathcal{L}_{\mathrm{Hungarian}}(\hat{P}^l_n,\hat{G}_n),
\end{equation}

\noindent where $\mathcal{D}^l(\cdot)$ denotes the $l$-th Transformer decoder layer in DETR. Note that in the default setting of our \textit{DataAug-DETR}, we use a set of object queries $Q$ shared with all images. During the Hungarian matching of different augmented versions of an image, the ground truth objects tend to be matched to different object queries.

\subsection{One-to-many Matching via Feature Augmentation}

In our $\textit{DataAug-DETR}$, data augmentations are applied to each image multiple times. The augmented $N$ versions of the same image are encoded by the feature backbone, whose computation cost is considerable as the $N$ augmented versions of each image need to be processed by the generally heavy backbone.

Since we choose to perform simple spatial transformations on each training image, the resulting features of the spatially transformed images can also be approximated by conducting the spatial transformations directly on the feature maps $F$ of each image $I$. In this way, each image only goes through the heavy feature backbone once and can still obtain multiple spatially augmented feature maps $\hat{F}_1,\dots,\hat{F}_N$. This strategy is much more efficient than \textit{DataAug-DETR} and we name it \textit{FeatAug-DETR}. The process can be formulated as:

\begin{equation}\label{eq:feat_aug1}
\begin{split}
&F=\mathcal{F}(I),\quad \hat{F}_n=\mathcal{E}(\mathcal{T}_n(F)), \\ 
&\hat{G}_n=\mathcal{T}_n(G),\quad \mathrm{for}\ n=1,2,\dots,N,
\end{split}
\end{equation}

\noindent where $\mathcal{E}(\cdot)$ denotes the Transformer encoder of DETR. $\mathcal{T}_n(\cdot)$ denotes conducting spatial augmentation on the feature map $F$. We perform feature augmentation to the output feature of the backbone and before the Transformer encoder $\mathcal{E}(\cdot)$, which we experimentally found to achieve better performance. The detailed experiment on the selection of the feature augmentation position can be found in Section \ref{sec:selection}.

After the augmented feature $\hat{F}_n$ is obtained, a matching process similar to that of \textit{DataAug-DETR} is applied:

\begin{equation}\label{eq:feat_aug2}
\hat{P}^l_n=\mathcal{D}^l(Q, \hat{F}_n), \ 
\mathcal{L}_{\mathrm{FeatAug}}=\sum_{n=1}^N\sum_{l=1}^L\mathcal{L}_{\mathrm{Hungarian}}(\hat{P}^l_n,\hat{G}_n),
\end{equation}

For the specific operation of feature augmentation $\mathcal{T}_n(\cdot)$, we investigate feature map flipping or/and cropping.

\subsubsection{Feature Map Flipping}
Horizontal flipping is performed on the feature map $F$ (denoted as \textit{FeatAug-Flip}), which is formulated as:

\begin{equation}\label{eq:feat_aug_flip}
\begin{split}
F=\mathcal{F}(I), \ &\hat{F}_1=\mathcal{E}(F),\ \hat{F}_2=\mathcal{E}(\mathrm{Flip}(F)), \\
&\hat{G}_1=G, \quad \ \ \hat{G}_2=\mathrm{Flip}(G).\\
\end{split}
\end{equation}

After applying \textit{FeatAug-Flip}, the two augmented feature maps $\hat{F}_1$ and $\hat{F}_2$ are forwarded to the follow-up modules of DETR and two separate Hungarian matchings are conducted for the two feature maps in the same training batch following Eq. \ref{eq:one2many}.

\subsubsection{Feature Map Cropping}
Besides the flip operation, random cropping on the feature map is tested (denoted as \textit{FeatAug-Crop}). The process of \textit{FeatAug-Crop} is similar to that of \textit{FeatAug-Flip} but replaces flipping with feature cropping.

The cropping and resizing hyperparameters are the same as the DETR's original image data augmentation cropping and resizing scheme. Since some state-of-the-art DETR-like frameworks are based on Deformable-DETR, which utilizes multi-scale features from a backbone, our \textit{FeatAug-Crop} augments both single-scale and multi-scale feature maps. Deformable-DETR utilizes multi-scale feature maps of 1/8, 1/16, and 1/32 original resolutions, respectively. In order to crop the features of the three scales, we use the RoIAlign \cite{2017maskrcnn} to crop and resize the same region across the three scales. The feature augmentation for single-scale features is the same but only conducts the augmentation on one scale. 

When cropping the features with RoIAlign, we find that the features after cropping $\hat{F}$ become blurry due to the bilinear interpolation in RoIAlign. It causes a domain gap between the original feature $F$ and the cropped feature $\hat{F}$. When DETR is trained with both types of region features, its detection performance deteriorates. We propose to use extra feature projectors on the cropped features to narrow down the domain gap. Given the three-scale features from the backbone, three individual projectors are adopted to transform the cropped features $\hat{F}$, each of which includes a $1\times1$ convolution layer and a group normalization layer \cite{2018group_norm}. The projectors are only applied on $\hat{F}$ during training, while the original features $F$ are still processed by the original detection head. The cropped feature projectors are able to mitigate the domain gap produced by RoIAlign and avoid performance reduction.

\subsubsection{Combining Flipping and Cropping}
In \textit{FeatAug-Flip} and \textit{FeatAug-Crop}, the two versions of feature maps of each image, which include the original and the augmented features, are forwarded through the DETR detection head. It is straightforward to combine the above flipping and cropping operations for feature augmentation. We name this combined version \textit{FeatAug-FC}. When applying \textit{FeatAug-FC}, three versions of each image's feature maps, i.e. the original, flipped, and cropped feature maps, are input into the DETR head.

\textit{DataAug-DETR} and \textit{FeatAug-DETR} are introduced to augment feature maps in the same training batch to realize one-to-many matching for DETRs from a new perspective. Both our methods improve detection performance and \textit{FeatAug-DETR} also significantly accelerates DETR training.

\section{Experiment} \label{sec:exp}

\subsection{Dataset and Implementation Details}\label{sec:detail}
The experiments are conducted on COCO 2017 object detection dataset \cite{2014coco}. The dataset is split into \texttt{train2017} and \texttt{val2017}, in which the \texttt{train2017} and \texttt{val2017} sets contain 118k and 5k images, respectively. There are 7 instances per image on average, up to 63 instances in a single image in the training set. We report the standard average precision (AP) results on COCO \texttt{val2017}. The DETR frameworks are tested with ResNet-50 \cite{2016resnet}, Swin-Tiny, and Swin-Large \cite{2021swin} backbones. The ResNet-50 and Swin-Tiny backbones are pretrained on ImageNet-1K \cite{2009imagenet} and Swin-Large is pretrained on ImageNet-22K \cite{2009imagenet}. 

We test our proposed methods on top of DAB-DETR \cite{2022dab_detr}, Deformable-DETR \cite{2021deformable_detr} with tricks (denoted as ``Deform-DETR w/ tck.'') from \cite{2022hybrdi_matching}, and Hybrid Matching Deformable-DETR ($\mathcal{H}$-Deformable-DETR) \cite{2022hybrdi_matching}. The latter two DETR frameworks use additional tricks, including bounding box refinement \cite{2021deformable_detr}, two-stage \cite{2021deformable_detr}, mixed query selection \cite{2022dino} and look forward twice \cite{2022dino}. These tricks accelerate the convergence speed and improve the final performance. 

We use the L1 loss and GIOU \cite{2019giou} loss for box regression, and focal loss \cite{2017focal} with $ 
\alpha\ =\ 0.25$, $\gamma\ =\ 2$ for classification. As the setting in DETR \cite{2020detr}, we apply auxiliary losses after each decoder layer. Similar to Deformable-DETR \cite{2021deformable_detr}, we add extra intermediate losses after the query selection module, with the same components as for each decoder layer. We adopt the loss coefficients: 2.0 for classification loss, 5.0 for L1 loss, and 2.0 for GIOU loss, which is the same as \cite{2022hybrdi_matching}.

\begin{table*}[tb!]
\caption{Main results of proposed data augmentation and feature augmentation on various DETR frameworks}\label{table:result}
  \centering
  \begin{threeparttable}
  \setlength{\tabcolsep}{2.8mm}{
  \begin{tabular}{lcccccccccccc}
    \toprule
    Method & Backbone & Tricks & \#queries & \#epochs &AP &AP$_{50}$ &AP$_{75}$ &AP$_S$ &AP$_M$ &AP$_L$ \\
    \midrule
    DETR-DC5 \cite{2020detr} & R50 & \XSolidBrush  & 300 & 500 & 43.3 & 63.1 & 45.9 & 22.5 & 47.3 & 61.1 \\
    Conditional-DETR-DC5 \cite{2021conditional_detr} & R50  & \XSolidBrush  & 300 & 108 & 45.1 & 65.4 & 48.5 & 25.3 & 49.0 & 62.2 \\
    DAB-DETR-DC5 \cite{2022dab_detr} & R50  & \XSolidBrush & 300 & 50 & 45.7 & 66.2 & 49.0 & 26.1 & 49.4 & 63.1 \\
    \rowcolor{gray!15} \ \ +\textit{FeatAug-FC} & R50  & \XSolidBrush & 300 & 36 & 46.1 ($\uparrow$0.4) & 66.6 & 49.5 & 27.7 & 50.0 & 62.8 \\
    \rowcolor{gray!15} \ \ +\textit{FeatAug-FC} & R50  & \XSolidBrush & 300 & 50 & 47.1 ($\uparrow$1.4) & 67.4 & 50.8 & 27.9 & 50.9 & 64.3 \\   
    \midrule
    Deformable-DETR-One-Stage \cite{2021deformable_detr} & R50 & \XSolidBrush & 300 & 50 & 44.5 & 63.6 & 48.7 & 27.1 & 47.6 & 59.6 \\
    Deformable-DETR \cite{2021deformable_detr} & R50 & \XSolidBrush & 300 & 50 & 46.9 & 65.6 & 51.0 & 29.6 & 50.1 &  61.6  \\
    DAB-Deformable-DETR \cite{2022dab_detr} & R50 & \XSolidBrush & 300 & 50 & 46.9 & 66.0 & 50.4 & 29.1 & 49.8 & 62.3  \\
    DN-Deformable-DETR \cite{2022dn_detr} & R50  & \XSolidBrush & 300 & 12 & 43.4 & 61.9 & 47.2 & 24.8 & 46.8 & 59.4  \\
    DN-Deformable-DETR \cite{2022dn_detr} & R50  & \XSolidBrush & 300 & 50 & 48.6 & 67.4 & 52.7 & 31.0 & 52.0 & 63.7  \\
    DINO-Deformable-DETR \cite{2022dino}$^\dag$ & R50  & \CheckmarkBold & 900 & 12 & 47.9 & 65.3 & 52.1 & 31.2 & 50.9 & 61.9 \\
    DINO-Deformable-DETR \cite{2022dino}$^\dag$ & R50  & \CheckmarkBold & 900 & 36 & 50.5 & 68.3 & 55.1 & 32.7 & 53.9 & 64.9 \\
    DINO-Deformable-DETR \cite{2022dino}$^\dag$ & Swin-L & \CheckmarkBold & 900 & 12 & 56.8 & 75.6 & 62.1 & 39.9 & 60.4 & 73.3 \\
    DINO-Deformable-DETR \cite{2022dino}$^\dag$ & Swin-L & \CheckmarkBold & 900 & 36 & 57.8 & 76.5 & 63.2 & 40.6 & 61.8 & 73.5 \\
    \midrule
    Deformable-DETR w/ tck. \cite{2021deformable_detr} & R50 & \CheckmarkBold & 300 & 12 & 47.0 & 65.3 & 50.9 & 30.7 & 50.0& 60.9 \\
    \rowcolor{gray!15} \ \ +\textit{FeatAug-FC}  & R50 & \CheckmarkBold & 300 & 12 & 48.7 ($\uparrow$1.7) & 67.1 & 53.3 & 31.2 & 52.1 & 63.2 \\
    Deformable-DETR w/ tck. \cite{2021deformable_detr} & R50 & \CheckmarkBold & 300 & 36 & 49.0 & 67.6 & 53.3 & 32.7 & 51.5 & 63.7 \\
    \rowcolor{purple!8} \ \ +\textit{DataAug}  & R50 & \CheckmarkBold & 300 & 48 & 50.0 ($\uparrow$1.0) & 68.6 & 54.6 & 33.3 & 52.9 & 64.5 \\
    \rowcolor{gray!15} \ \ +\textit{FeatAug-FC}  & R50  & \CheckmarkBold & 300 & 24 & 49.9 ($\uparrow$0.9) & 68.3 & 54.7 & 32.5 & 52.9 & 65.1 \\
    Deformable-DETR w/ tck. \cite{2021deformable_detr} & Swin-T & \CheckmarkBold & 300 & 12 & 49.3 & 67.8 & 53.4 & 31.6 & 52.4 & 64.4  \\
    \rowcolor{gray!15} \ \ +\textit{FeatAug-FC}  & Swin-T  & \CheckmarkBold & 300 & 12 & 50.9 ($\uparrow$1.6) & 69.4 & 55.6 & 33.1 & 54.3 & 66.1 \\
    Deformable-DETR w/ tck. \cite{2021deformable_detr} & Swin-T & \CheckmarkBold & 300 & 36 & 51.8 & 70.9 & 56.5 & 34.6 & 55.1 & 67.9  \\
    \rowcolor{purple!8} \ \ +\textit{DataAug}  & Swin-T & \CheckmarkBold & 300 & 48 & 53.3 ($\uparrow$1.5) & 72.1 & 58.5 & 35.9 & 56.7 & 68.4 \\
    \rowcolor{gray!15} \ \ +\textit{FeatAug-FC}  & Swin-T & \CheckmarkBold & 300 & 24 & 52.7 ($\uparrow$0.9) & 71.3 & 57.5 & 35.1 & 56.6 & 67.8 \\
    Deformable-DETR w/ tck. \cite{2021deformable_detr} & Swin-L & \CheckmarkBold & 300 & 12 & 54.5 & 74.0 & 59.2 & 37.0 & 58.6 & 71.0  \\
    \rowcolor{gray!15} \ \ +\textit{FeatAug-FC}  & Swin-L & \CheckmarkBold & 300 & 12 & 55.8 ($\uparrow$1.3) & 75.5 & 61.1 & 39.2 & 60.2 & 72.1 \\
    Deformable-DETR w/ tck. \cite{2021deformable_detr} & Swin-L & \CheckmarkBold & 300 & 36 & 56.3 & 75.7 & 61.5 & 39.1 & 60.3 & 71.9  \\
    \rowcolor{purple!8} \ \ +\textit{DataAug}  & Swin-L & \CheckmarkBold & 300 & 48 & 57.0 ($\uparrow$0.7) & 76.3 & 62.2 & 40.4 & 60.9 & 73.3 \\
    \rowcolor{gray!15} \ \ +\textit{FeatAug-FC}  & Swin-L & \CheckmarkBold & 300 & 24 & 57.1 ($\uparrow$0.8) & 76.4 & 62.5 & 41.1 & 60.9 & 73.3 \\
    \rowcolor{gray!15} \ \ +\textit{FeatAug-FC}$^\dag$  & Swin-L  & \CheckmarkBold & 900 & 24 & 57.6 & 76.7 & 63.1 & 41.1 & 61.5 & 73.8 \\
    \midrule
    $\mathcal{H}$-Deformable-DETR \cite{2022hybrdi_matching} & R50  & \CheckmarkBold & 300 & 12 & 48.7 & 66.7 & 53.4 & 31.4 & 51.8 & 63.6 \\
    \rowcolor{gray!15} \ \ +\textit{FeatAug-Flip} & R50 & \CheckmarkBold & 300 & 12 & 49.4 ($\uparrow$0.7) & 67.5 & 53.7 & 31.9 & 52.7 & 64.3 \\
    $\mathcal{H}$-Deformable-DETR \cite{2022hybrdi_matching} & R50 & \CheckmarkBold  & 300 & 36 & 50.0 & 68.1 & 54.6 & 32.5 & 53.1 & 65.0 \\
    \rowcolor{gray!15} \ \ +\textit{FeatAug-Flip} & R50 & \CheckmarkBold & 300 & 24 & 50.4 ($\uparrow$0.4) & 68.7 & 54.9 & 32.7 & 53.6 & 65.2 \\
    $\mathcal{H}$-Deformable-DETR \cite{2022hybrdi_matching} & Swin-L & \CheckmarkBold  & 300 & 12 & 55.9 & 75.1 & 61.0 & 39.3 & 59.9 & 72.1  \\
    \rowcolor{gray!15} \ \ +\textit{FeatAug-Flip} & Swin-L & \CheckmarkBold  & 300 & 12 & 56.4 ($\uparrow$0.5) & 75.6 & 61.8 & 40.0 & 60.4 & 72.4 \\
    $\mathcal{H}$-Deformable-DETR \cite{2022hybrdi_matching} & Swin-L & \CheckmarkBold  & 300 & 36 & 57.1 & 76.3 & 62.7 & 40.2 & 61.6 & 73.3  \\
    \rowcolor{gray!15} \ \ +\textit{FeatAug-Flip}  & Swin-L & \CheckmarkBold  & 300 & 24 & 57.6 ($\uparrow$0.5) & 76.6 & 63.1 & 40.4 & 61.9 & 73.9 \\
    $\mathcal{H}$-Deformable-DETR \cite{2022hybrdi_matching}$^\dag$ & Swin-L & \CheckmarkBold  & 900 & 36 & 57.9 & 76.9 & 63.8 & 42.5& 62.0 & 73.5  \\
    \rowcolor{gray!15} \ \ +\textit{FeatAug-Flip}$^\dag$ & Swin-L & \CheckmarkBold  & 900 & 24 & \textbf{58.3} ($\uparrow$0.4) & 77.1 & 64.0 & 41.7 & 62.4 & 73.9 \\
    \bottomrule
  \end{tabular}}
  \begin{tablenotes}
      \small
      \item Tricks: tricks described in Section \ref{sec:detail}. $^\dag$: keep 300 instead of 100 predictions for evaluation.
    \end{tablenotes}
  \end{threeparttable}
\end{table*}

Each tested DETR framework is composed of a feature backbone, a Transformer encoder, a Transformer decoder, and two prediction heads for boxes and labels. All Transformer weights are initialized with Xavier initialization \cite{2010xavier}. In the experiments, we use 6 layers for both the Transformer encoder and decoder. The hidden dimension of the Transformer layers is 256. The intermediate size of the feed-forward layers in the Transformer blocks is 2048, which follows the settings of \cite{2022hybrdi_matching}. The MLP networks for box and label predictions share the same parameters across different decoder layers. We use 300 object queries in the decoder. We use AdamW \cite{2019adamw} optimizer with a weight decay $10^{-4}$. We use an initial learning rate of $2\times 10^{-4}$ for the Deformable DETR head and a learning rate of $2\times 10^{-5}$ for the backbone, which is the same as those in \cite{2021deformable_detr}. A $1/10$ learning rate drop is applied at the 11-th, 20-th, and 30-th epochs for the 12, 24, and 36 epoch settings, respectively. The model is trained without dropout. The training batch size is 16 and the experiments are run on 16 NVIDIA V100 GPUs. During validation, we select 100 predicted boxes and labels with the largest classification logits for evaluation by default.

\subsection{Main Results}
Our \textit{DataAug-DETR} and \textit{FeatAug-DETR} methods are compatible with most DETR-like frameworks. The results on COCO \texttt{val2017} set are shown in Table \ref{table:result}. The DAB-DETR-DC5-\textit{FeatAug-FC} denotes applying \textit{FeatAug-FC} on top of DAB-DETR \cite{2022dab_detr} with R50 dilated C5-stage image features \cite{2017FCN}, Deformable-DETR w/ tck. -\textit{FeatAug-FC} denotes \textit{FeatAug-FC} on top of Deformable-DETR with tricks, and $\mathcal{H}$-Deformable-DETR-\textit{FeatAug-Flip} denotes \textit{FeatAug-Flip} on top of Hybrid matching Deformable-DETR. We experimentally found that, when integrating with Hybrid Matching, \textit{FeatAug-Flip} achieves better performance than \textit{FeatAug-FC} as shown in Table \ref{table:hybrid_compare}. Thus, we report the results of Hybrid Matching with \textit{FeatAug-Flip} and others with \textit{FeatAug-FC}.

\begin{table}[tb!]
\centering
\caption{\textit{FeatAug-FC} vs. \textit{FeatAug-Flip} on top of Hybrid Matching Deformable-DETR}\label{table:hybrid_compare}
  \setlength{\tabcolsep}{1.7mm}{
  \begin{tabular}{@{}lcccccccc@{}}
    \toprule
    Method & Backbone & \#epochs &AP &AP$_{50}$ &AP$_{75}$ \\
    \midrule
    $\mathcal{H}$-Deformable-DETR \cite{2022hybrdi_matching} & R50 & 24 & 50.0 & 68.1 & 54.6 \\
    \ \ +\textit{FeatAug-FC} & R50 & 24 & 50.1 & 68.2 & 54.6 \\
    \ \ +\textit{FeatAug-Flip} & R50 & 24 & 50.4 & 68.7 & 54.9 \\
    \bottomrule
  \end{tabular}}
\end{table}

In the table, we also report the performances of some previous representative DETR variants. The compared methods include single-scale and multi-scale detectors. The Deformable-DETR ones, which utilize multi-scale features, generally achieve better performances than single-scale detectors. 

We compare the performance gain of our \textit{DataAug-DETR} and \textit{FeatAug-DETR} on top of the baselines. \textit{DataAug-DETR} is evaluated on top of Deform-DETR w/ tck. \cite{2021deformable_detr}. \textit{FeatAug-DETR} is evaluated on DAB-DETR \cite{2022dab_detr}, Deform-DETR w/ tck., and $\mathcal{H}$-Deformable-DETR \cite{2022hybrdi_matching}. \textit{DataAug-DETR} (48 epochs training) improves the performance by about 1.0 AP compared with the Deform-DETR w/ tck. (36 epochs training) baseline, while the performance of the baseline trained with 48 epochs degrades. In each epoch of \textit{DataAug-DETR}, we only train $1/N$ of the whole training set, where $N$ is the augmentation times of \textit{DataAug-DETR}. Thus, the training time per epoch of \textit{DataAug-DETR} is the same as that of the baseline. The detailed investigation of the convergence speed of \textit{DataAug-DETR} is shown in Section \ref{sec:dataaug_speed}. 

On the single-scale DAB-DETR with ResNet-50 backbone, \textit{FeatAug-FC} improves the performance for 1.4 AP with 50 training epochs and reaches 47.1 AP, which makes the single-scale detector's performance better than the multi-scale Deformable-DETR \cite{2021deformable_detr} (46.9 AP). It is shown that the AP$_M$ and AP$_L$ of DAB-DETR-DC5-\textit{FeatAug-FC} exceed those of Deformable-DETR by large margins (0.8 AP and 2.7 AP, respectively). When shortening the training epochs to 36 epochs, \textit{FeatAug-FC} still achieves the performance of 46.1 AP, which is 0.4 AP higher than the DAB-DETR-DC5 baseline trained with 50 epochs. 

On top of the multi-scale Deform-DETR w/ tck. baseline, \textit{FeatAug-FC} not only achieves around 1.0 AP gain after convergence but also shortens the training epochs (24 epochs vs. 36 epochs) on all the three tested backbones. Our \textit{FeatAug-FC} integrated into Deform-DETR w/ tck. on Swin-Large backbone achieves 57.6 AP, which is comparable with the state-of-the-art DINO-Deformable-DETR framework.

\begin{figure*}[tb!]
    \centering
    \includegraphics[width=0.495\linewidth]{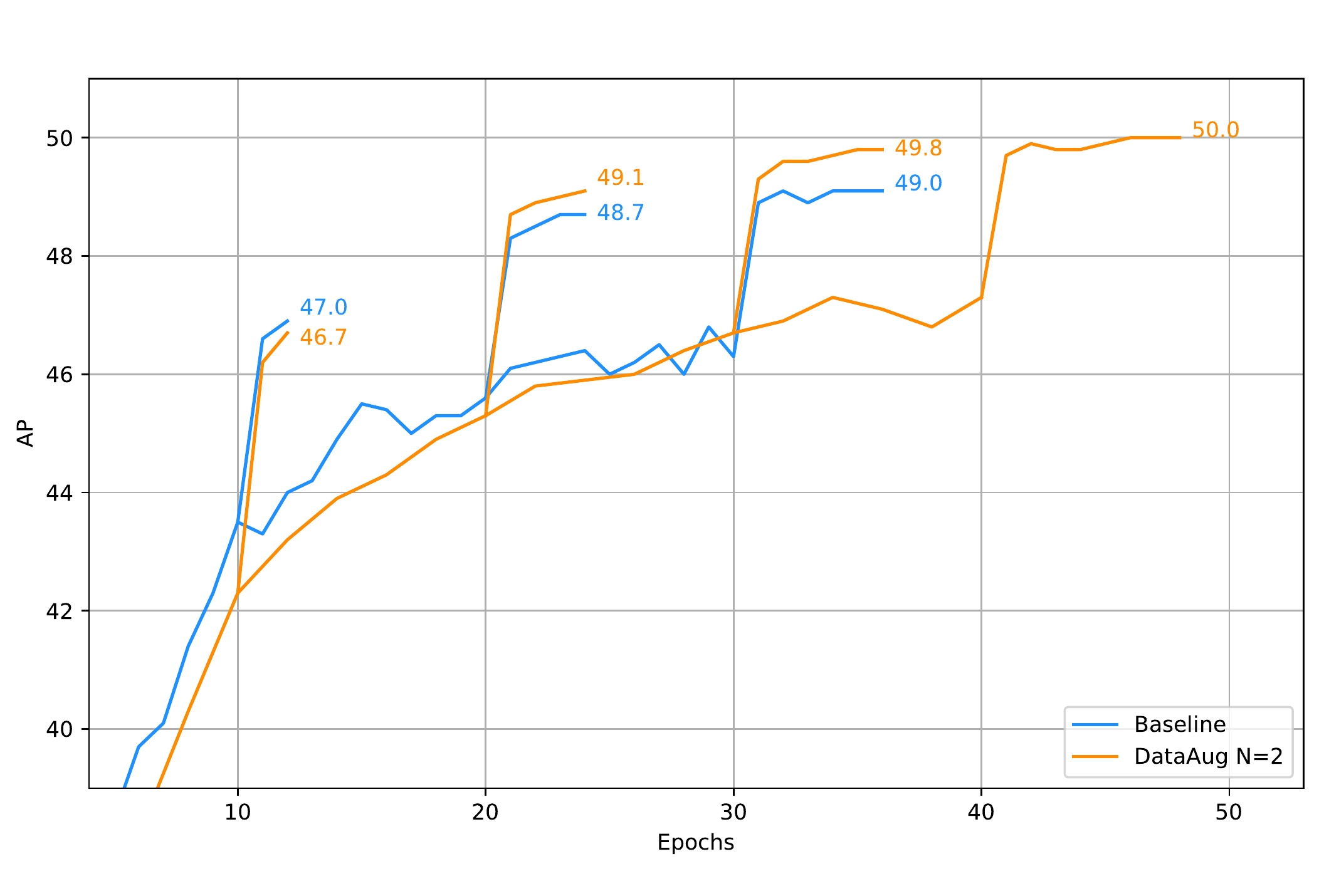}
    \includegraphics[width=0.495\linewidth]{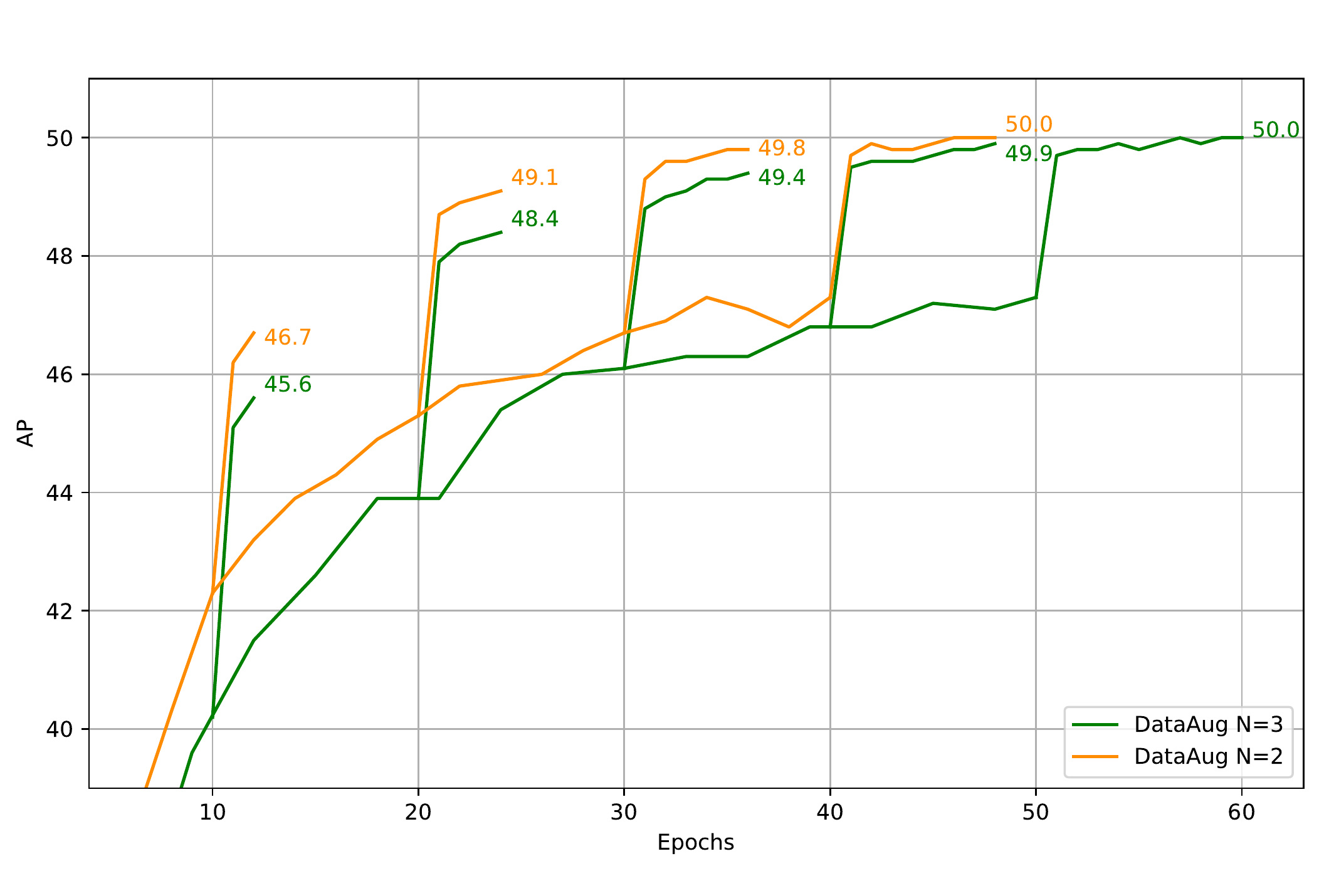}
    \caption{\textbf{Left:} Training AP curves of Deform-DETR w/ tck. baseline and \textit{DataAug-DETR} with $N=2$. \textbf{Right:} Training AP curves of \textit{DataAug-DETR} with $N=2$ and $N=3$. It shows that augmenting each image for more than two times results in slower convergence and is non-beneficial to the final performance.}
    \label{fig:converge}
\end{figure*}

\begin{table}[tb!]
\caption{The performance of \textit{DataAug-DETR} with non-spatial transformation} \label{table:non_spatial}
\centering
\setlength{\tabcolsep}{3.5mm}{
\begin{tabular}{lcccc}
\toprule
\multirow{2}{*}{Method} & \multirow{2}{*}{Backbone} & \multicolumn{3}{c}{\#epochs} \\
\cmidrule{3-5}
&& 12 & 24 & 36 \\
\midrule
Deform-DETR w/ tck. & R50 & 47.0 & 48.7 & 49.0 \\
\ \ +\textit{DataAug} & R50 & 46.7 & 49.1 & 49.8 \\
\ \ +\textit{DataAug-Resizing} & R50 & 45.2 & 48.1 & 48.9\\
\bottomrule
\end{tabular}}
\end{table}

The work Hybrid Matching (denoted as $\mathcal{H}$) \cite{2022hybrdi_matching} also tackles the one-to-many matching problem and achieves state-of-the-art performance. We also test our \textit{FeatAug-DETR} on top of Hybrid Matching with Swin-Large backbone and achieve the performance of 58.3 AP, which is 0.5 AP higher than the previous state-of-the-art DINO. In experiments, we adopt the default hyperparameter settings of Hybrid Matching as \cite{2022hybrdi_matching}.

\subsection{Non-spatial vs. Spatial Transformation for \textit{DataAug-DETR}}
As discussed in Section \ref{sec:pilot}, the object queries' assignments with the ground truths are sensitive to position changes. In other words, when an augmentation changes the relative position of objects, these objects almost always match different queries compared to the original image/feature. Our proposed flipping and cropping augmentations are both spatial transformations that change relative object positions. There are also other widely used data augmentation methods that do not change the objects' relative positions, such as image random resizing. In this section, we also test applying only image random resizing in our \textit{DataAug-DETR} method, which applies random resizing of each image several times in the same batch. We compare the non-spatial transformation with the Deform-DETR w/ tck. baseline and our \textit{DataAug-DETR} with default settings. In the experiments, the augmentation times for each image $N=2$.

As shown in Table \ref{table:non_spatial}, the performance of only applying image resizing in \textit{DataAug-DETR} is even worse than the baseline and our proposed default \textit{DataAug-DETR} setting. The model converges slower and leads to worse performance than the baseline. This experiment shows that spatial transformations such as flipping and cropping are crucial in the effectiveness of \textit{DataAug-DETR}. It also shows the rationality of the flipping and cropping feature augmentation in \textit{FeatAug-DETR}.

\begin{table}[tb!]
\caption{Convergence analysis of \textit{DataAug-DETR}} \label{table:converge}
\centering
\setlength{\tabcolsep}{2.6mm}{
\begin{tabular}{lccccc}
\toprule
\multirow{2}{*}{Method} & \multirow{2}{*}{Backbone} & \multicolumn{4}{c}{\#epochs} \\
\cmidrule{3-6}
&& 12 & 24 & 36 & 48 \\
\midrule
Deform-DETR w/ tck. & R50 & 47.0 & 48.7 & \textbf{49.0} & 48.3\\
\ \ +\textit{DataAug}-$N=2$ & R50 & 46.7 & 49.1 & 49.8 & \textbf{50.0}\\
\ \ +\textit{DataAug}-$N=3$ & R50 & 45.6 & 48.4 & 49.4 & \textbf{49.9}\\
\bottomrule
\end{tabular}}
\end{table}

\subsection{Convergence Analysis of \textit{DataAug-DETR}} \label{sec:dataaug_speed}

In the following analysis experiments, unless otherwise specified, we test our proposed method and evaluate its different designs on top of the Deform-DETR w/ tck. and ResNet-50 backbone and treat it as our experiment baseline. 

When keeping the same $16$ batch size as the baseline, \textit{DataAug-DETR} can be viewed as changing the order of training data compared with the ordinary training pipeline. Here we investigate the convergence speed of \textit{DataAug-DETR} AP-epoch curves.

As shown in Figure \ref{fig:converge} (left) and Table \ref{table:converge}, applying \textit{DataAug-DETR} improves the final converged performance. The baseline converges with 36 epochs and achieves 49.0 AP. Further training hurts the model as the performance drops when trained for 48 epochs. After applying \textit{DataAug-DETR} with augmentation times $N=2$, its performance at 36 epochs is 0.8 AP higher than the baseline. The performance continues to improve with a 48-epoch training scheme. It is also observed that \textit{DataAug-DETR} slightly slows down convergence in early epochs.

We also investigate the augmentation number $N$ with \textit{DataAug-DETR}, Figure \ref{fig:converge} (right) shows that the convergence is slower with a larger augmentation number, while the final performance becomes saturated after $N\geq2$. Thus we adopt $N=2$ as the default setting unless otherwise specified.

\subsection{Comparison of Different Feature Augmentation Operators}
We compare the performance of the proposed \textit{FeatAug-Flip}, \textit{FeatAug-Crop}, and \textit{FeatAug-FC} on the Deform-DETR w/ tck. baseline. The results are listed in Table \ref{table:feataug}.

As shown in the table, all of our \textit{FeatAug-DETR}'s results are better than that of the baseline. \textit{FeatAug-Flip} is 0.3 AP higher than \textit{FeatAug-Crop} with 12 training epochs. The stronger \textit{FeatAug-FC} is 0.3 AP better than \textit{FeatAug-Flip} and \textit{FeatAug-Crop} with 24 epochs and reaches 49.9 AP. 

\begin{table}[tb!]
	\caption{Results of different \textit{FeatAug-DETR} methods on Deform-DETR w/ tck.}\label{table:feataug}
	\centering
	\setlength{\tabcolsep}{1.7mm}{
		\begin{tabular}{@{}lccccccc@{}}
			\toprule
			Method & \#ep. &AP &AP$_{50}$ &AP$_{75}$ &AP$_S$ &AP$_M$ &AP$_L$ \\
			\midrule
			Deform-DETR w/ tck.\hspace{-5pt} & 12 & 47.0 & 65.3 & 50.9 & 30.7 & 50.0& 60.9 \\
			Deform-DETR w/ tck.\hspace{-5pt} & 36 & 49.0 & 67.6 & 53.3 & 32.7 & 51.5 & 63.7 \\
			\midrule
			\ \ +\textit{FeatAug-Flip} \hspace{-5pt} & 12 & 48.4 & 66.7 & 52.8 & 31.7 & 51.4 & 64.0  \\
			\ \ +\textit{FeatAug-Flip} \hspace{-5pt}  & 24 & 49.6 & 68.1 & 54.1 & 31.6 & 53.1 & 64.8 \\
			\ \ +\textit{FeatAug-Crop} \hspace{-5pt} & 12 & 48.1 & 66.3 & 52.3 & 30.7 & 51.3 & 63.2 \\
			\ \ +\textit{FeatAug-Crop} \hspace{-5pt}  & 24 & 49.6 & 68.1 & 54.1 & 33.2 & 52.6 & 64.7 \\
			\ \ +\textit{FeatAug-FC} & 12 & 48.7 & 67.1 & 53.3 & 31.2 & 52.1 & 63.2 \\
			\ \ +\textit{FeatAug-FC}  & 24 & \textbf{49.9} & 68.3 & 54.7 & 32.5 & 52.9 & 65.1 \\
			\bottomrule
	\end{tabular}}
\end{table}

\begin{table}[tb!]
\caption{Computational Consumption of \textit{FeatAug-DETR} on Deform-DETR w/ tck. with 16 NVIDIA V100 GPUs} \label{table:computation}
\centering
\setlength{\tabcolsep}{2.1mm}{
\begin{tabular}{@{}lccc@{}}
\toprule
 Method & Backbone & Time (per epoch) & Time (total) \\
\midrule
Deform-DETR w/ tck. & R50 & 40min & 24h (36epochs) \\
\ \ +\textit{FeatAug-Flip} & R50 & 60min & 24h (24epochs) \\
\midrule
Deform-DETR w/ tck. & Swin-T & 50min & 30h (36epochs) \\
\ \ +\textit{FeatAug-Flip} & Swin-T & 70min & 28h (24epochs) \\
\midrule
Deform-DETR w/ tck. & Swin-L & 140min &  84h (36epochs)\\
\ \ +\textit{FeatAug-Flip} & Swin-L & 160min & 64h (24epochs) \\
\ \ +\textit{FeatAug-FC} & Swin-L & 180min & 72h (24epochs) \\
\bottomrule
\end{tabular}}
\end{table}

The results show that \textit{FeatAug-Flip} converges faster than \textit{FeatAug-Crop}, while their convergence performance is similar. \textit{FeatAug-FC} provides further performance improvement compared with \textit{FeatAug-Flip} and \textit{FeatAug-Crop}, while it also requires extra training time for epoch. Thus, \textit{FeatAug-Flip} is suitable for models with relatively small backbones (e.g. R50 and Swin-Tiny). \textit{FeatAug-FC} is preferred when training with large backbones (e.g. Swin-Large).

\begin{figure*}[tb!]
  \centering
   \includegraphics[width=0.495\linewidth]{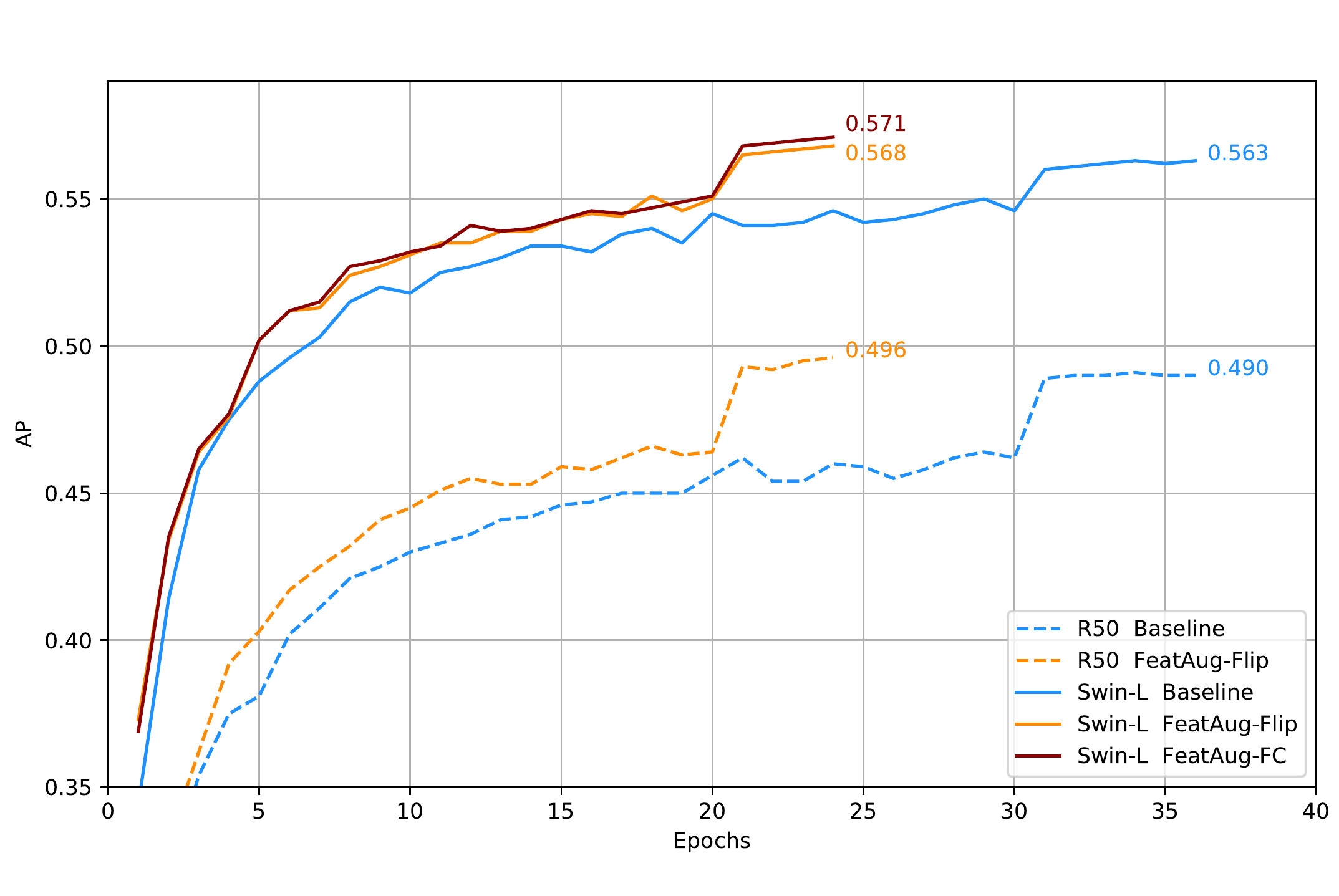}
   \includegraphics[width=0.495\linewidth]{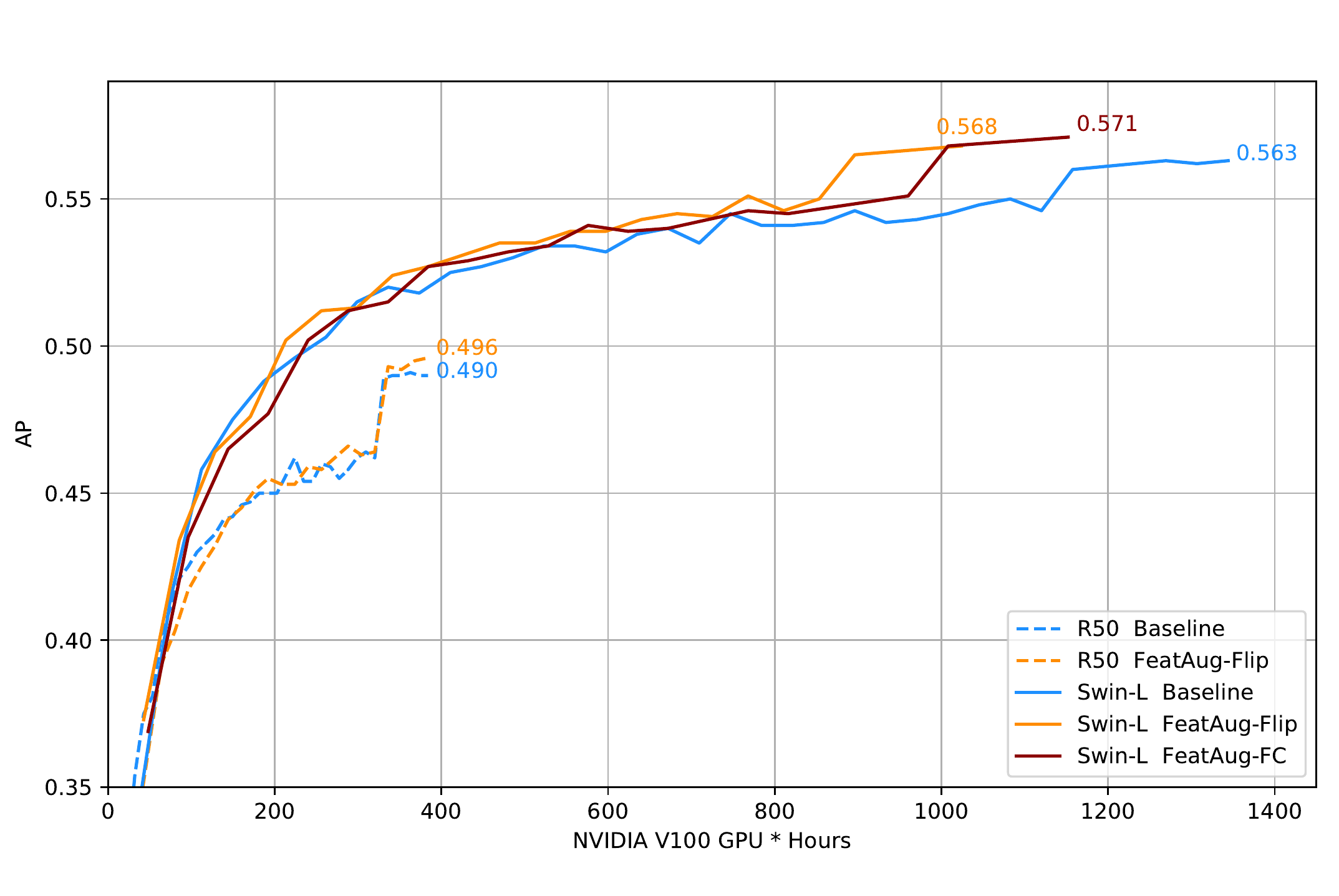}
   \caption{\textbf{Left: }Training AP curves w.r.t training epochs of Deform-DETR w/ tck. and that integrated with our \textit{FeatAug-DETR}. \textbf{Right: }Traing AP curves w.r.t GPU$\times$Hours of Deform-DETR w/ tck. and that integrated with our \textit{FeatAug-DETR}.}
   \label{fig:feataug_converge}
\end{figure*}

\subsection{Convergence Analysis of \textit{FeatAug-DETR}}
In \textit{FeatAug-DETR}, the Transformer encoder and decoder process several augmented features during training, which produce extra computational consumption. However, such an increase in computation is invariant to the scale of the backbone. When training with large-scale backbones, the extra computational consumption is relatively small compared with the backbone. The training time per epoch comparisons with different backbones is shown in Table \ref{table:computation}.

\begin{table}[tb!]
\centering
\caption{Evaluation on One-stage Deformable-DETR}\label{table:one_stage}
  \setlength{\tabcolsep}{2.2mm}{
  \begin{tabular}{@{}lccccccc@{}}
    \toprule
    Method & \#ep. &AP &AP$_{50}$ &AP$_{75}$ &AP$_S$ &AP$_M$ &AP$_L$ \\
    \midrule
    One-stage & 50 & 44.5 & 63.6 & 48.7 & 27.1 & 47.6 & 59.6 \\
    \ \ +\textit{FeatAug-FC} & 36 & 45.9 & 64.8 & 50.3 & 27.7 & 49.3 & 60.6 \\
    \ \ +\textit{FeatAug-FC}  & 50 & \textbf{46.3} & 65.3 & 50.6 & 28.0 & 49.7 & 61.1 \\
    \bottomrule
  \end{tabular}}
\end{table}

\begin{table}[tb!]
\centering
\caption{Ablation on input feature selection in \textit{FeatAug-DETR}}\label{table:selection}
  \setlength{\tabcolsep}{3.6mm}{
  \begin{tabular}{lccccccc}
    \toprule
    Method & \#ep. &AP &AP$_{50}$ &AP$_{75}$ \\
    \midrule
    Deform-DETR w/ tck. & 12 & 47.0 & 65.3 & 50.9 \\
    \ \ +\textit{FeatAug-Flip-Encoder} & 12 & 46.5 & 64.8 & 50.6 \\
    \ \ +\textit{FeatAug-Flip-Default}  & 12 & \textbf{48.4} & 66.7 & 52.8  \\
    \bottomrule
  \end{tabular}}
\end{table}

\begin{table}[tb!]
\centering
\caption{Ablation on cropped feature projector in \textit{FeatAug-Crop} on Deform-DETR w/ tck.}\label{table:projector}
  \setlength{\tabcolsep}{1.6mm}{
  \begin{tabular}{@{}lccccccc@{}}
    \toprule
    Method & \#ep. &AP &AP$_{50}$ &AP$_{75}$ &AP$_S$ &AP$_M$ &AP$_L$ \\
    \midrule
    Deform-DETR w/ tck. & 36 & 49.0 & 67.6 & 53.3 & 32.7 & 51.5 & 63.7 \\
    Same Proj. & 24 & 49.2 & 67.8 & 53.6 & 31.8 & 52.9 & 63.4 \\
    Cropped Proj. & 24 & \textbf{49.6} & 68.1 & 54.1 & 33.2 & 52.6 & 64.7  \\
    \bottomrule
  \end{tabular}}
\end{table}

The table reports the training time of the Deform-DETR w/ tck. baseline and that integrated with \textit{FeatAug-Flip}. The training time of \textit{FeatAug-Crop} is similar to \textit{FeatAug-Flip}. Our training process uses 16 NVIDIA V100 GPUs with a batch size of 16. As shown in Table \ref{table:computation}, the increase in the computation time is invariant with the size of the backbone. The proportion of the increased computation time is only about 15\% when using the Swin-L backbone.

The training AP curves of \textit{FeatAug-DETR} are shown in Figure \ref{fig:feataug_converge}. The left figure shows the curves w.r.t. training epochs. Our \textit{FeatAug-DETR} consistently surpasses the Deform-DETR w/ tck. baseline by a large margin and shortens the training epochs from 36 epochs to 24 epochs. Considering our \textit{FeatAug-DETR} requires extra computation consumption in each epoch, we also show the training AP curves w.r.t. GPU$\times$Hours in the right figure. It shows \textit{FeatAug-DETR} still achieves better performance than the baseline in terms of detection accuracy and convergence speed. Especially when the backbone scale is larger, the performance gain is more obvious.

\subsection{Ablation Studies}
\subsubsection{Evaluation on one-stage Deformable-DETR}
Since we mainly evaluate our method based on Deform-DETR w/ tck., which consists of tricks that boost performance. In order to show that our methods are independent of these tricks and can still improve various DETR variants' performance, we also test our method on top of the one-stage Deformable-DETR, which is its original version. The results are shown in Table \ref{table:one_stage}.

Our method accelerates its convergence speed. \textit{FeatAug-FC} trained for 36 epochs is 1.4 AP better than the one-stage Deformable-DETR trained for 50 epochs. With the same 50 training epochs, \textit{FeatAug-FC} reaches 46.3 AP, which is 1.8 AP better than the baseline.

\subsubsection{Input feature of FeatAug-DETR}\label{sec:selection}
In our \textit{FeatAug-DETR}, we augment the image feature from the backbone and input the augmented features into the Transformer encoder. However, in the Deform-DETR w/ tck. baseline, the features after the Transformer encoder can also be chosen as the alternative for augmentation. To analyze which feature is better as the input of \textit{FeatAug-DETR}, we test them and the results are shown in Table \ref{table:selection}. \textit{FeatAug-DETR} on the feature after the Transformer encoder is denoted as \textit{FeatAug-Encoder} and the augmentation on the feature directly from the backbone is denoted as \textit{FeatAug-Default}.

\begin{figure}[tb!]
    \centering
    \includegraphics[width=0.32\linewidth]{fig/img.jpg}
    \includegraphics[width=0.32\linewidth]{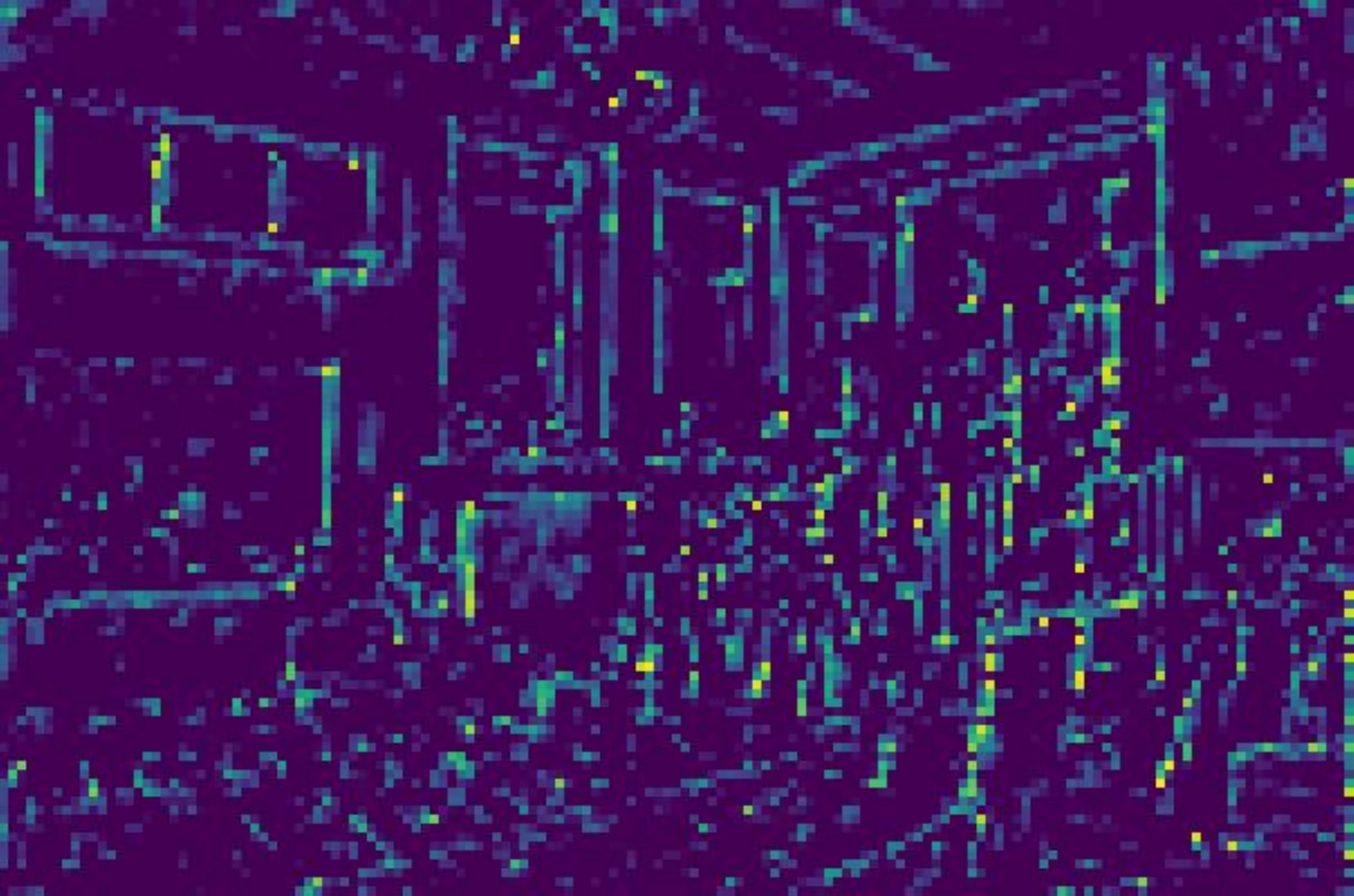}
    \includegraphics[width=0.32\linewidth]{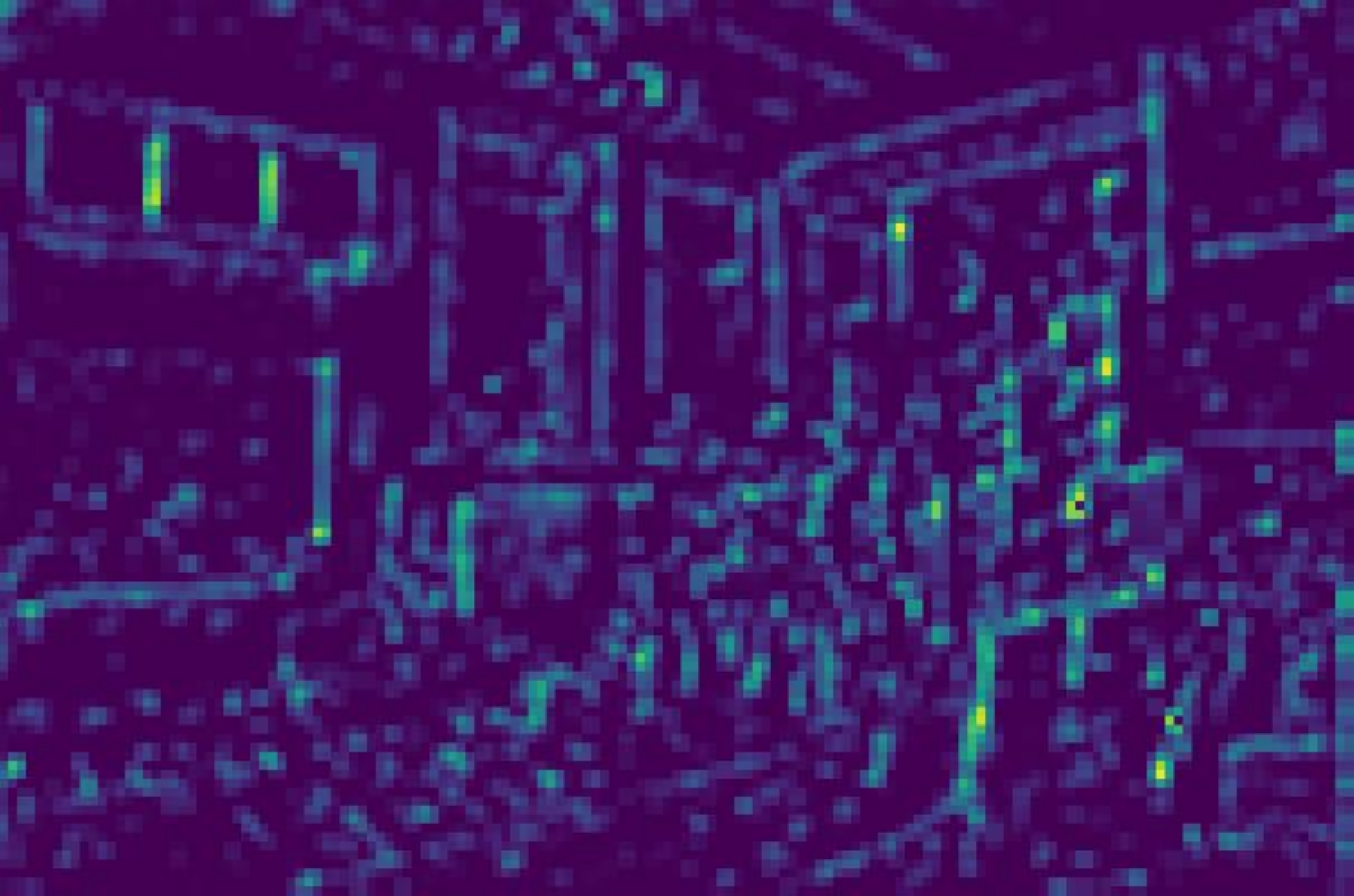}
    \caption{Visualization of randomly selected one channel feature of the output of vision backbone. It visualizes the feature domain gap caused by RoIAlign, we adopt projectors after the cropped features to mitigate the domain gap. \textbf{Left:} Original input image. \textbf{Middle:} Original feature from vision backbone. \textbf{Right:} Feature after RoIAlign operation. It shows that the augmented feature becomes blurry compared with the original one, and such a domain gap causes detection performance degradation. Here we adopt the $1/8$ input-resolution feature map from the R50 backbone trained with Deform-DETR w/ tck.}
    \label{fig:domain_gap}
\end{figure}

As shown in the table, the feature augmentation on the encoder feature actually hurts the performance and is even worse than that of the baseline. Since the Transformer encoder enhances the image features with positional information. If the feature augmentation is applied after the Transformer encoder, it would mislead the later Transformer decoder about the positional information of this image and thus harm the Transformer decoder training.

\subsubsection{Cropped feature projectors in \textit{FeatAug-DETR}}
In our \textit{FeatAug-Crop} and \textit{FeatAug-FC}, in order to mitigate the domain gap caused by RoIAlign, we adopt projectors after the cropped features. The visualization of the caused domain gap is shown in Figure \ref{fig:domain_gap}. Here we ablate on removing the individual projectors and using the same projector on both the original and cropped features. The results are shown in Table \ref{table:projector}. Using the projectors for cropped features in \textit{FeatAug-Crop} performs 0.4 AP better than using the same projector for both cropped and original features. The performance gain is more obvious on small objects and large objects, where $AP_S$ increases by 1.4 and $AP_L$ by 1.3. 

\begin{figure}[tb!]
  \centering
   \includegraphics[width=1\linewidth]{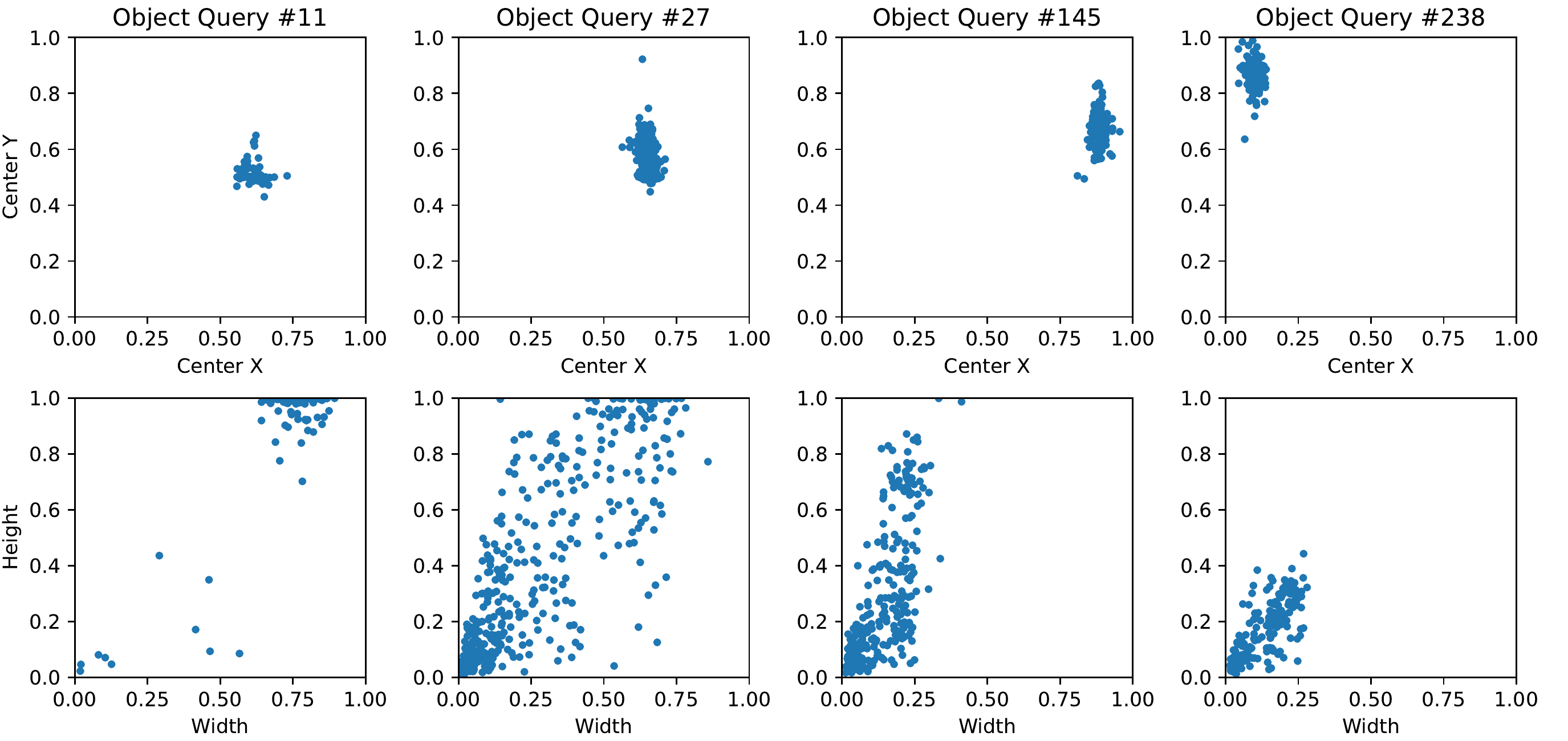}
   \caption{We visualize the predicted bounding boxes position distribution of 4 randomly selected object queries on 10000 random selected images from COCO \texttt{train2017} set. The top 4 figures show the predicted center x and center y of the corresponding bounding boxes, while the bottom 4 figures show the predicted heights and widths of the queries.}
   \label{fig:analysis}
\end{figure}

\subsection{Visualization of Position-Aware Object Queries}\label{sec:pos_aware}

In the pilot study in Section \ref{sec:motivation}, it shows that spatial augmentation changes the query-object assignments. We further visualize some bounding boxes that each object query predicts. Here we test a trained Deformable-DETR without our \textit{DataAug-DETR} or \textit{FeatAug-DETR}. We record all the output bounding box predictions that are matched with the ground truth objects for each query and visualize $\{$center\_x, center\_y, width, height$\}$ of all the predicted bounding boxes for each query. Here we visualize four queries (\#11, \#27, \#145, and \#238) in Figure \ref{fig:analysis}.

As shown in Figure \ref{fig:analysis}, the predicted center of each object query is almost always located around a fixed position, which means that a spatial transformation that changes the bounding boxes' relative positions on the feature map (such as flipping) can change the matched object queries. Furthermore, by observing the predicted heights and widths distribution of Query \#11 and Query \#27, it shows that though the two queries predict similar center positions, their predicted height and width are differently distributed. Query \#11 always predicts large objects while Query \#27 is always in charge of small objects. It shows that the cropping operation that changes the sizes of the predicted bounding boxes' also varies the object-query matchings. 

The above two observations show that the object queries assigned to predicted bounding boxes are highly position-aware. Applying flipping and cropping operations makes different object queries match different ground truth objects.

\section{Conclusion}
In this paper, we enrich the formulation of one-to-many matching for DETRs. We summarize the one-to-many matching mechanism of augmenting object queries for Group-DETR \cite{2022group_detr} and Hybrid Matching \cite{2022hybrdi_matching} and propose augmenting image features to implement one-to-many matching. To be detailed, we proposed \textit{DataAug-DETR} method to help DETR-like methods to achieve higher performance after convergence. Further, we also propose the \textit{FeatAug-DETR} method, including \textit{FeatAug-Flip}. \textit{FeatAug-Crop}, and \textit{FeatAug-FC}. The \textit{FeatAug-DETR} method significantly accelerates DETR training and accomplishes better performance. \textit{FeatAug-DETR} improve the performance of Deformable-DETR \cite{2021deformable_detr} with different backbones for around 1.0 AP and shorten the training epochs to only $1\times$ or $2\times$ standard training schedule. When applying \textit{FeatAug-DETR} together with Hybrid Matching \cite{2022hybrdi_matching} and using a Swin-Large backbone, we achieve the current state-of-the-art performance of 58.3 AP.

{\small
\bibliographystyle{IEEEtran}
\bibliography{egbib}
}

\vfill

\end{document}